\documentclass[a4paper, 10pt, conference]{ieeeconf}     

\usepackage{FG2017}
\usepackage{times}
\usepackage{epsfig}
\usepackage{graphicx}
\usepackage{amsmath}
\usepackage{amssymb}
\usepackage{url}
\usepackage{graphicx}
\usepackage{float}
\usepackage{caption}
\graphicspath{ {images/} }

\FGfinalcopy 

\IEEEoverridecommandlockouts                             
\title{\LARGE \bf
Predicting First Impressions with Deep Learning
}

\author{\parbox{16cm}{\centering
    {\large Mel McCurrie$^1$,  Fernando Beletti$^1$, Lucas Parzianello$^1$, Allen Westendorp$^1$,\\ Samuel Anthony$^{2,3}$, and Walter J. Scheirer$^1$}\bigbreak
    {\normalsize
    $^1$Department of Computer Science and Engineering, University of Notre Dame\\
    $^2$Department of Psychology, Harvard University 
    $^3$Perceptive Automata, Inc.    
    }}
}

\begin{document}
\IEEEoverridecommandlockouts\pubid{\makebox[\columnwidth]{978-1-5090-4023-0/17/\$31.00~\copyright{}2017 IEEE \hfill} \hspace{\columnsep}\makebox[\columnwidth]{ }}

\ifFGfinal
\thispagestyle{empty}
\pagestyle{empty}
\else

\pagestyle{plain}
\fi
\maketitle

\begin{abstract}
Describable visual facial attributes are now commonplace in human biometrics and affective computing, with existing algorithms even reaching a sufficient point of maturity for placement into commercial products. These algorithms model objective facets of facial appearance, such as hair and eye color, expression, and aspects of the geometry of the face. A natural extension, which has not been studied to any great extent thus far, is the ability to model subjective attributes that are assigned to a face based purely on visual judgements. For instance, with just a glance, our first impression of a face may lead us to believe that a person is smart, worthy of our trust, and perhaps even our admiration --- regardless of the underlying truth behind such attributes. Psychologists believe that these judgements are based on a variety of factors such as emotional states, personality traits, and other physiognomic cues. But work in this direction leads to an interesting question: how do we create models for problems where there is no ground truth, only measurable behavior? In this paper, we introduce a convolutional neural network-based regression framework that allows  us  to  train  predictive  models  of  crowd  behavior for social attribute assignment. Over images from the AFLW face database, these models demonstrate strong correlations with human crowd ratings.   

\end{abstract}

\begin{figure}[t!]
\centering
\includegraphics[width=8cm]{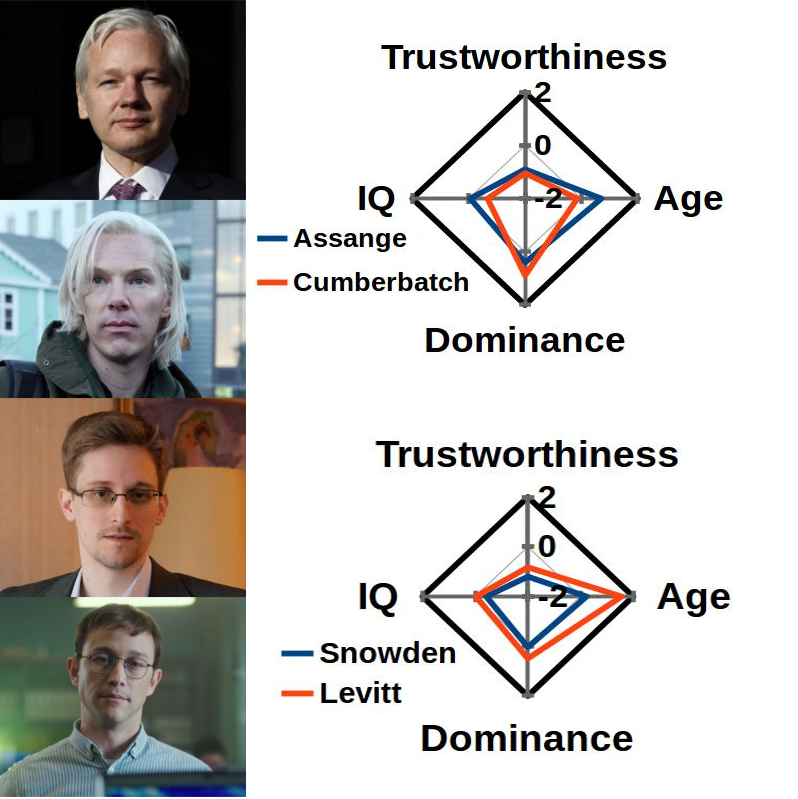}
\caption[LoF entry]{Computational modeling of social attributes allows us to predict what the crowd might say about a face image.  In this image we graphically compare the attribute predictions for Julian Assange and Benedict Cumberbatch, who plays Assange in the movie \textit{The Fifth Estate}, as well as the predictions for Edward Snowden and Joseph Gordon-Levitt, who plays Snowden in the movie \textit{Snowden}. Specifically looking at these images, our models output similar predictions between the subjects and their actors.
%attesting to the accuracy of the portrayals in the films. 
The radar plots above reflect the output of a face processing pipeline, where faces are detected, aligned, and then processed through a deep convolutional neural network regressor that models a particular social attribute. This regression framework is the main contribution of our work. For this image we display the predictions' z-scores with respect to the training data.
}
\vspace{-3mm}
\label{fig:teaser}
\end{figure}

\section{Introduction}

In human attribute modeling there often exists a disparity between the way humans describe humans and the way computational models describe humans. A large amount of describable attribute research in computer vision concentrates on objective traits. For example, work using the popular CelebA dataset~\cite{liu2015deep, rudd2016moon, zhong2016face, zhang2016gender} applies different methods to model traits such as ``Male" and ``Bearded" with binary annotations. Beyond objective attributes, it is possible to model more subjective traits such as expression~\cite{graves2008facial, dumas2001emotional}, attractiveness~\cite{kumar2011describable}, and humorousness~\cite{lewenberg2016predicting}, but research often overlooks the important interrelation between attribute modeling and social psychology. Enabling computers to make accurate predictions about objective content and enabling computers to make human-like judgements about subjective content are both necessary steps in the development of machine intelligence. Here we focus on the latter.

Specifically, we concentrate on descriptions of the face, as an abundance of social psychology research demonstrates a human tendency to make judgements in social interactions based on the faces of fellow humans~\cite{senior1999investigation, winston2002automatic, Alicke01}. Popular human characteristics of academic interest closely related to these social interactions include emotion~\cite{mignault2003many}, attractiveness~\cite{Alicke01}, trustworthiness~\cite{todorov2008reading, winston2002automatic, pinkham2008investigation, falvello2015robustness}, dominance~\cite{senior1999investigation, mignault2003many}, sociability, intelligence, and morality~\cite{Alicke01}. Psychologists often specifically concentrate on trustworthiness, dominance, and intelligence because they represent comprehensive abstract qualities that humans regard in each other. Alexander Todorov, one of the foremost psychologists studying these social judgements uses dominance and trustworthiness as the basis of many in-depth studies of human judgement~\cite{todorov2008reading, todorov2009evaluating, todorov2008evaluating}. Ultimately he finds that most other recognizable subjective traits in humans can be represented as an orthogonal function of dominance and trustworthiness~\cite{oosterhof2008functional}, which suggests these two conceptual traits are ideal for computational modeling. 

Closely related to our work is research concentrated on the assessment of abstract traits in human faces based on the effect of facial contortions and positions. Inspired by animals' displays of dominance and submissiveness in respective head raises and bows, Mignault et al. specifically analyzed the effects of head tilt on the change in perceived dominance and emotion~\cite{mignault2003many}. Not only does the study confirm the hypothesized disparity in perceived traits based on head tilt, but it also finds gender has a noteworthy influence on subjects' perceptions. Keating et al. assessed the effect of eyebrow and mouth gestures on perceived dominance and happiness in a cultural context~\cite{keating1981culture}. The study found smiling to be a universal indicator of happiness and showed weak associations between not smiling and dominance. It also determined the effect of a lowered-brow on perceived dominance to be generally restricted to Western subjects. 

In this paper we connect traditional machine learning and social psychology findings like those described above. We work specifically with traits that do not have a ground truth and can be considered abstract representations of high-level human attributes. Additionally, we introduce a convolutional neural network-based (CNN) regression framework that allows us to train predictive models of crowd behavior for social attribute assignment. Very different from prior work, we make use of a unique visual psychophysics crowdsourcing platform, TestMyBrain.org, to gather the annotations necessary for training. As a case study, we examine three purely (when analyzed in a visual context) social attributes: dominance, trustworthiness, and IQ. We also look at the more familiar objective attribute of age, but purely in the context of crowd judgements. Our models demonstrate strong correlations with crowd ratings, which we suspect are largely driven by low-level image queues.

In short, our contributions in this paper are:
\begin{itemize}
  \item{A novel ground truth-free dataset of over 6,000 images annotated for all four traits of interest.} 
  \item{The deployment of a crowd-sourced data collection regime, which collects large amounts of data on high-level social attributes from the popular psychophysics testing platform TestMyBrain.org.}
  \item{The comparison of different deep learning architectures for abstract social attribute modeling.}
  \item{A set of highly effective automatic predictors of social attributes that have not been modeled before in computer vision.}
\end{itemize}

\begin{figure*}[t!]
    \centering
    \includegraphics[width=17cm,height=7cm]{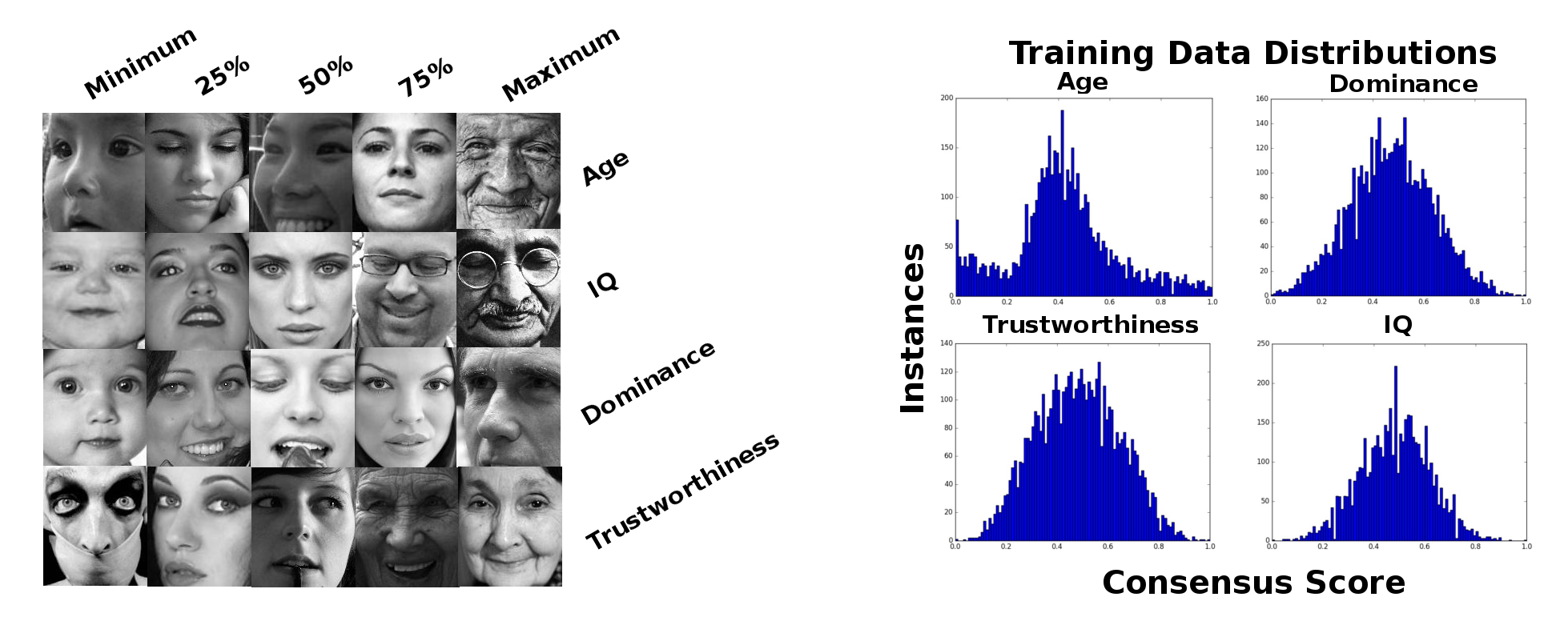}
    \caption{We assert that to most accurately model humans' psychological judgements, each of these traits should be modeled on a continuous distribution. For this reason we employed the Likert Scale in our data collection and then took the average of human ratings for each image. This graphic shows faces at each quartile from the dataset  (left) as well as the training data distributions (right), all of which seem to be close to normal.}  
    \label{fig:quartilesanddistributions}
\end{figure*}

\section{Related Work}

The related work in computer vision falls into two categories: general facial attributes, and specific CNN-based approaches. We review both in this section.

\subsection{Attributes in Computer Vision}

Due to the proliferation of low-cost high performance computing resources (\textit{e.g.}, GPUs) and web-scale image data, large-scale image classification and labeling is now commonplace in computer vision. With respect to face images from the web, Labeled Faces in the Wild~\cite{Huang2012a}, YouTube Faces~\cite{wolf2011face}, MegaFace~\cite{2016mf2},  Janus Benchmark A~\cite{Klare_2015_CVPR}, and CelebA~\cite{liu2015deep} are all popular choices for a variety of facial modeling tasks beyond conventional face recognition. Attribute prediction, where the objective is to assign semantically meaningful labels to faces in order to build a human interpretable description of facial appearance, is the particular task we concentrate on in this paper.

Both Farhadi et al.~\cite{farhadi2009describing} and Lampert et al.~\cite{lampert2009learning} originally conceived of visual attributes as a development supporting object recognition, rather than a primary goal in and of itself. Faces, however, are a special case where standalone analysis supports applications in biometrics and affective computing.
 Kumar et al. used facial attributes for face verification and image search~\cite{kumar2011describable}. Scheirer et al.  applied the statistical extreme value theory to facial attribute search spaces to create accurate multi-dimensional representations of attribute searches~\cite{scheirer2012multi}. Siddiquie et al. modeled the relationships between different attributes to create more accurate multi-attribute searches~\cite{siddiquie2011image}. And Luo et al. captured the interdependencies of local face regions to increase classification accuracy~\cite{luo2013deep}.

Certain traits such as Age~\cite{montillo2009age, kwon1999age, levi2015age} and gender~\cite{lewenberg2016predicting, levi2015age} have enjoyed disproportionate attention, but researchers also model numerous other facial attributes. The release of the large CelebA dataset~\cite{liu2015deep} also prompted several novel studies of facial attributes on all $40$ traits in the dataset~\cite{rudd2016moon, zhong2016face, zhang2016gender}. For a comprehensive review of facial attribute work in practical biometric systems, see the review authored by Dantcheva et al.~\cite{dantcheva2016else}.

\subsection{Convolutional Neural Networks for Attributes}

Current state-of-the-art facial attribute modeling relies on CNNs. Pioneering work in the field, Golomb et al. trained a CNN with an $8.1\%$ error rate on gender prediction~\cite{golomb1990sexnet}. More recently, Zhang et al. used CNNs alongside conventional part-based models to predict attributes such as clothing style, gender, action, and hair style from images~\cite{zhang2014panda}. Wang et al. applied CNNs to an automatically generated egocentric dataset annotated for contextual information such as weather and location~\cite{wang2016walk}. Levi et al. used a CNN for age and gender classification from faces~\cite{levi2015age}. Liu et al. used two cascaded CNNs and trained support vector machines to separate the processes of face localization and attribute prediction~\cite{liu2015deep}. And Zhong et al. extended the work of Liu et al. using off-the-shelf CNNs to build facial descriptors in a different approach to attribute prediction~\cite{zhong2016face}.

Most similar to our research is the recent work of Lewenberg et al.~\cite{lewenberg2016predicting}. They use a CNN to predict objective traits including gender, ethnicity, age, make-up, and hair color, and subjective traits including emotional state, attractiveness, and humorousness. That research introduced a new face attributes dataset of $10,000$ images annotated for these traits. To generate this dataset, Lewenberg et al. employed Amazon's Mechanical Turk raters from the US and Canada to rate a subset of the PubFig dataset, aggregating labels from three separate individuals for each image. Notably, the work only analyzes the traits with binary classification, labeling each image as ``yes" or ``no" with respect to a trait. Our most immediate improvement on this work is in the way in which we collect data. We use an online psychophysics testing platform, aggregating data from a larger number of raters from an arguably more reliable and geographically variable source. In addition, we model more abstract, representational traits on continuous distributions. 

Also parallel to our work, and the current state-of-the-art attribute prediction, is the work of Rudd et al.~\cite{rudd2016moon}.  Rudd et al. employ a single custom Mixed Objective Optimization Network (MOON) to multi-task facial attribute recognition, minimizing the error of their networks over all forty traits of the CelebA dataset~\cite{liu2015deep}. We use our own implementation of the MOON architecture as a basis for each separate trait in our modeling.

\section{Crowd-Sourced Data Collection}

In this paper we introduce a new dataset for social attribute modeling. The dataset consists of $6,300$ grayscale images of faces sampled from the AFLW dataset~\cite{tugraz:icg:lrs:koestinger11b} and annotated for the four traits we study. Representative samples of the dataset for each trait can be seen in Fig.~\ref{fig:quartilesanddistributions}. This dataset is novel in that there is no ground truth.  For traits such as Age and IQ, which are easy to record and described on well-known scales, it is of course possible to produce a dataset with verifiable ground truth annotations --- but this is not our objective. Rather than analyze and model actual trustworthiness, dominance, IQ, and age, we choose to study people's described perceptions of the aforementioned traits. For example, our dataset does not include actual ages, instead the images are annotated by a consensus score --- aggregate statistics of what many people said about the ages of the subjects in the images. 
\subsection{TestMyBrain.org}

For this high-level, ground truth-free annotation, we use TestMyBrain.org~\cite{germine2012web}, a crowd-sourced psychophysics testing website where users go to test and compare their mental abilities and preferences. It is one of the most popular ``brain testing" sites on the web, with over 1.6 million participants since 2008. But what specific advantages does TestMyBrain.org have over a service like Amazon's Mechanical Turk? 

TestMyBrain.org is a citizen science effort that facilitates psychological experiments and provides personalized feedback for the user, mutually benefiting both researchers and those curious about their own mind.  The subject pool is geographically diverse and provides an arguably superior psychometric testing group compared to smaller more homogeneous subject pools such as that of Mechanical Turk. In addition to being an ideal setting for aggregate, cross-cultural psychometric experiments for researchers, TestMyBrain.org provides the non-monetary incentive of detailed, personalized results for subjects. Subjects visiting the site are motivated by a desire to learn about themselves and have little incentive to respond to experiments quickly or poorly. Based on these factors, we determined that the subject pool of TestMyBrain.org is ideal for the delicate task of honestly appraising abstract, ground truth-free attributes in faces.

\begin{figure}[t]
\centering
\includegraphics[width=7cm]{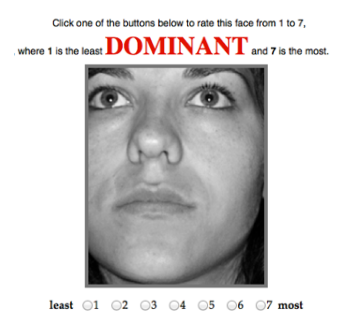}
\caption{A sample behavioral task that a subject might see on TestMyBrain.org. All ratings collected for this work were on a Likert scale between $1$-$7$, where $1$ indicates the least amount of attribute presence, and the $7$ indicates the most amount.}
\label{fig:tmb}
\end{figure}

Using TestMyBrain.org, we asked participants to judge faces for a select trait on a Likert Scale, a psychometric bipolar scaling method shown in Fig.~\ref{fig:tmb}. As can be seen in Table~\ref{tab:stats}, each face has an average of about 32 judgements for Trustworthiness and Dominance and 15 for Age and IQ. We recorded the average judgement to use as the consensus score for that image and normalized the Trustworthiness and Dominance scores. In training we map all $y$ values so that $0 \le y \le 1$. We calculated the coefficient of determination ($R^2$) of mean human ratings from two independent sets of $943$ subjects for $389$ random images from the AFLW set for Trustworthiness and $400$ random images from the AFLW set for dominance. The Trustworthiness $R^2$ is $0.93$ and the Dominance $R^2$ is $0.88$. Both of these statistics are very similar to the internal reliability calculated by Oosterhof and Todorov~\cite{oosterhof2008functional}. Thus there is indeed signal in these data that can be learned by a machine learning algorithm.

\section{CNN Regression for Social Attributes}

Our algorithm is a regression model that outputs a single score from an input image.  A regression, rather than a binary classification, is a more realistic representation of the initial judgements humans make. For example, from our four modeled traits, both Age and IQ are already known to be described by continuous distributions and are therefore likely judged on continuous distributions. We assert the other two modeled traits, Trustworthiness and Dominance, are similarly best described by continuous distributions. For what is discussed below with respect to architectures, assume the output is always a single floating point number from the fully-connected layers of a CNN after the convolutional layers' feature extraction.

\subsection{Comparing Architectures: What Works Best for Social Attribute Modeling?}

We initially compared five  architectures with conceptually similar structures but different depths and use of regularization. We ran each with similar parameters which we determined empirically. To test very deep architectures we used the Oxford Visual Geometry Group's VGG networks~\cite{simonyan2014very}. We reproduced VGGNet19's convolutional architecture, modifying the shape of the input and output matrices for our smaller grayscale images and single floating point regression output. To compare results from another deep, yet slightly more shallow architecture, we also modified and used VGGNet16 in the same manner.  

The newest architecture we analyzed is our implementation of the MOON architecture~\cite{rudd2016moon}, which is more shallow than both of the VGGNet implementations. The convolutional feature extracting portion of the architecture is similar to the VGG networks in that it consists of several segments, where each segment has multiple convolutional layers followed by a max pooling layer. We modify the architecture for our smaller grayscale images and connect the convolutional layers to fully-connected layers that output a single score. With respect to our reimplementation of the MOON architecture, we made use of the Keras~\cite{chollet2015keras} and Theano~\cite{bergstra2010theano} deep learning frameworks. 

For comparison we also added two shallower custom architectures with fewer convolutional layers and varying regularization. In the ``Shallow" network we employ three segments of convolution and max pooling connected to fully-connected layers with dropout and Parametric ReLU activations. In the ``Basic6" network we employ four segments of a single convolution and max pooling followed by two fully-connected layers with ReLU activations and no dropout.  

As will be discussed below in Sec.~\ref{sec:experiments}, the differences in model performances on the validation sets during training are not very large, suggesting the architecture choice may not make a significant difference. The newer MOON architecture performs slightly better on most of the traits, however, so we chose to use it as a basis for our final optimized models. Note that the earlier work of Lewenberg et al.~\cite{lewenberg2016predicting} used an AlexNet block structure augmented with supervised features (facial landmark information) and a custom loss function, while MOON is a more straightforward VGG~\cite{simonyan2014very} modification.

\begin{table}
\begin{center}
\caption{Statistics on the $5,040$ images used for training for all four social attribute classes (normalized to a $[0,1]$ range). The ``Mean Std. of Ratings" refers to the average standard deviation of the human scores for each individual image.}
 \label{tab:stats}
 \begin{tabular}{|| c c c c c ||}
 \hline
  & \textbf{Trust.} & \textbf{Dom.} & \textbf{Age} & \textbf{IQ} \\ [0.3ex] 
 \hline\hline
 \textbf{Mean of Ratings} & $0.48$ & $0.47$ & $0.42$ & $0.48$ \\
 \hline
 \textbf{Std. of Ratings} & $0.16$ & $0.16$ & $.20$ & $0.14$ \\
 \hline
 \textbf{Mean Std.} & & & & \\ \textbf{of Ratings} & $0.34$ & $0.32$ & $0.13$ & $0.27$  \\
 \hline
 \textbf{Mean Num.} & & & & \\  \textbf{of Ratings} & $32.47$ & $32.19$ & $15.80$ & $15.79$\\
 \hline
\end{tabular}
\end{center}
\end{table}

\begin{figure*}[t]
    \centering
    \includegraphics[width=\textwidth]{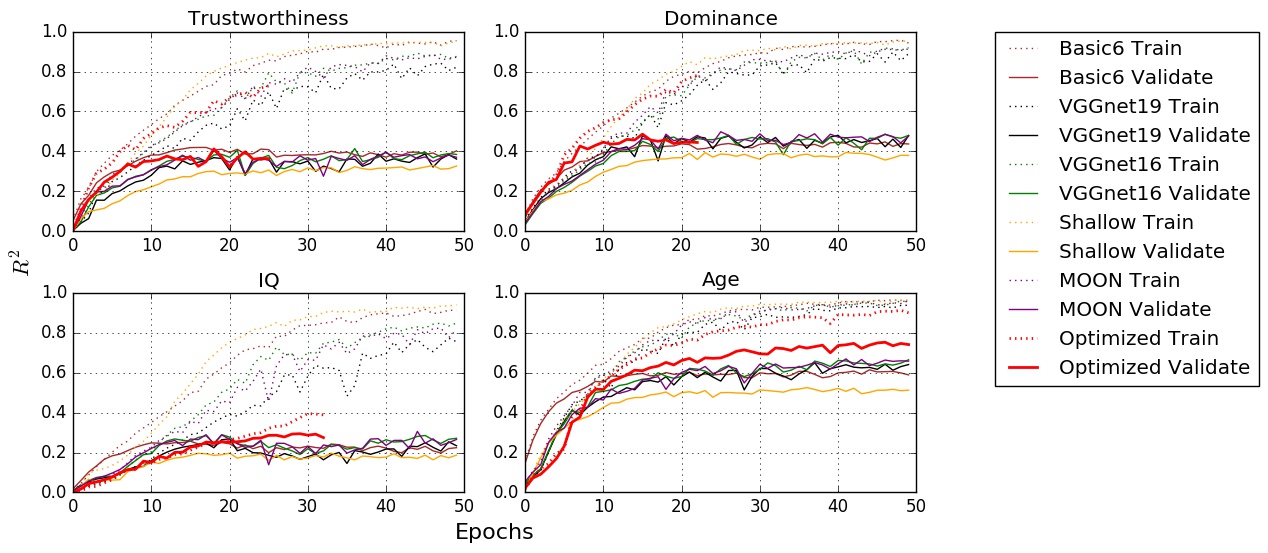}
    \caption{In this image we compare the ability of each architecture to learn the dataset and generalize to validation data. We include all four traits, training and validation scores, and all four original architectures (best viewed in color) plus our final architecture based on the hyperparameter optimization results. Of the four original architectures the MOON models generally perform better, but our optimized models consistently perform the best. (Optimized models were trained with early stopping, as can be seen in the plots.) }  
    \label{fig:trainhyperopt}
\end{figure*}

\subsection{Hyperparameter Optimization}

Rudd et al. train their MOON models on RGB images that are larger than our grayscale images and model hypothetically less abstract objective attributes from the CelebA dataset, which is annotated for binary classification. This suggests that our very different dataset and features could benefit from some deviations in parameter choices. 

To determine the best network size and deviation in parameters from the original MOON architecture, we optimize the network for each trait using hyperopt~\cite{bergstra2013hyperopt}, a python library for hyperparameter optimization. Our search space includes learning rate, dropout, the number of filters in each layer, the number of layers, the amount of data augmentation, and the parameters of a sampling function. Employing hyperopt with the Tree of Parzen Estimators (TPE) algorithm allowed us to test a multitude of different parameter and architecture combinations. After a very wide parameter search, we perform a refined search with early stopping, and use the best models.

We maximize the model's performance with respect to the 
%$R^2$ (or $min(1-R^2)$) where $R^2$ is the 
coefficient of determination ($R^2$) from the regression of $\hat{y}$, the model's predicted scores, on $y$, the average human annotations. We use the coefficient of determination as the measure of performance because it represents the percentage of prediction variation explained by the regression model. As explained previously, our measure of performance cannot be described as ``accuracy", as there is no ground truth.

As seen in Fig.~\ref{fig:trainhyperopt}, each trait trains very differently. Following this trend, each trait's coefficient of determination is optimized by slightly different hyperparameters and deviations from the MOON architecture as seen in Table \ref{tab:hyperparameters}. However the improvements are only modest, suggesting that deeper architectures and data augmentation are not helpful for this task.

\section{Experimental Evaluation}
\label{sec:experiments}

There are two important facets of evaluation with respect to our social attribute models: (1) model correlation with human crowd ratings of images, and (2) feature importance for social attribute models. Each of these facets is explored in this section. After data collection, our dataset consisted of $6,300$ grayscale images of faces, aligned to correct for in-plane rotation using the CSU Toolkit~\cite{bolme2003csu} and annotated for Trustworthiness, Dominance, Age, and IQ. We randomly separated  $80\%$ of the original dataset into a training set, and split the remaining $1260$ images into a validation and test set (630 images each). The test set is held out during training, while the validation set is used to tune the hyperparameters.

\subsection{Correlations with Crowd Judgements}

We employ the $R^2$ value from a regression of $\hat{y}$, the model's predicted scores, on $y$, the original human annotations, as a measure of our model's performance. As seen in the supplemental material, this is a reasonable metric given the linear relationship of $y$ and $\hat{y}$. To properly compare architectures and assess the training performance of our optimized model, we record the $R^2$ at each epoch and graph them in Fig.~\ref{fig:trainhyperopt}. 

Looking at the graphs, the validation $R^2$ values are ultimately very similar between architectures. There is some variability in training speed, and randomly good weight initializations seem to help, however the depth of the architecture does not seem to explain any improvement in scores.

As expected, our optimized architectures outperform the other four architectures. We display our final results from the optimized networks in Table \ref{tab:rsquared}, which shows $R^2$ values from regressions of our model's predicted values on human annotated consensus scores for both the validation and testing sets.  Each trait has a slightly different coefficient of determination, however all scores are strong for a psychology-oriented experiment incorporating noisy human measurements. Our models for Age are the strongest, IQ are the weakest, and Trustworthiness and Dominance perform similarly to each other.

\begin{figure}[t]
    \centering
    \includegraphics[width=6.9cm]{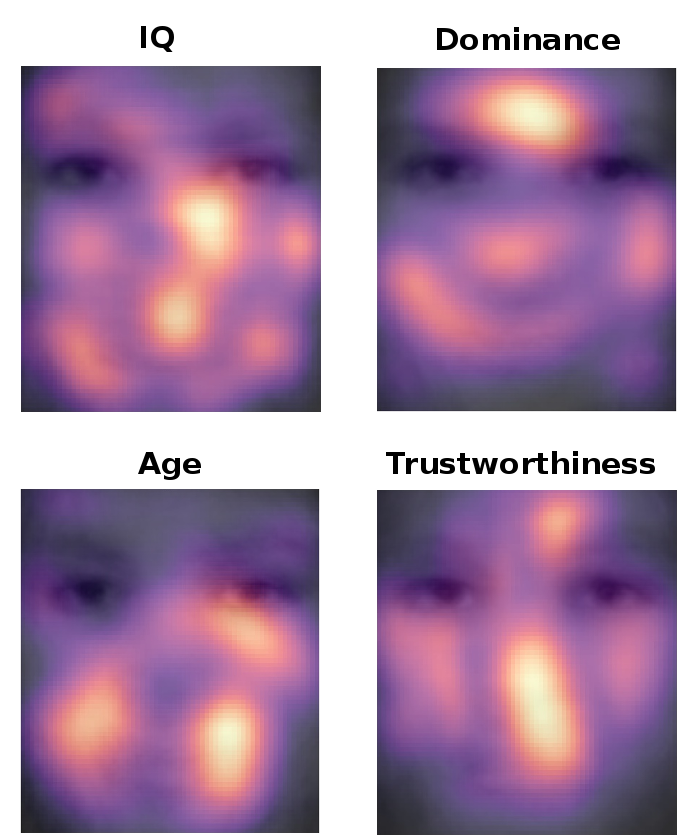}
    \caption{We can visualize regions of the face that are most important to the trained models by systematically covering parts of the face and recording the absolute differences. Here we separately analyze 100 images of the validation set and display the average differences as a heatmap on top of the averaged faces.}
    \label{fig:averageheatmaps}
\end{figure}

\begin{figure}[t]
    \centering
    \includegraphics[width=9cm]{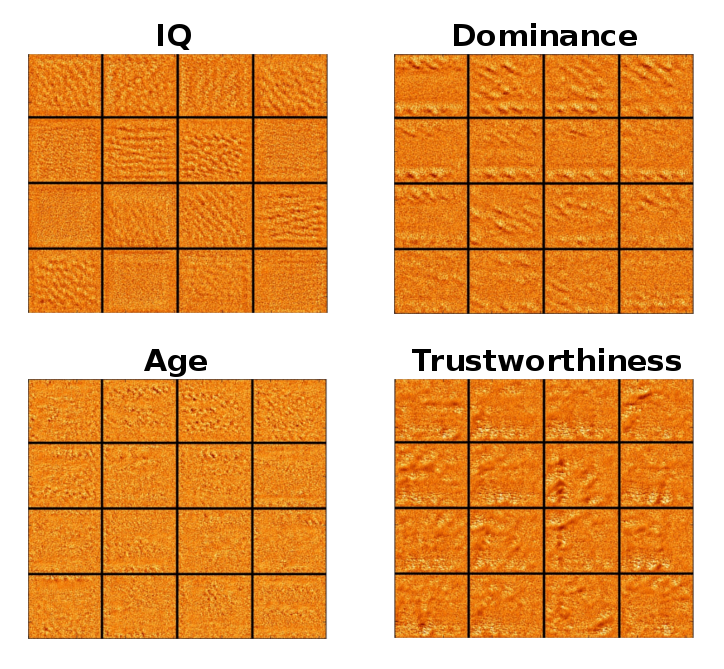}
    \caption{Visualizing a sample of the filters from the last convolutional layer of each optimized model, we can observe the resemblance of the output to a low-level feature extractor, consistent with our observation that deeper architectures add little to no improvement. (Color added to improve contrast.)}
    \label{fig:filters}
\end{figure}

\begin{figure*}[t]
    \centering
    \includegraphics[width=\textwidth]{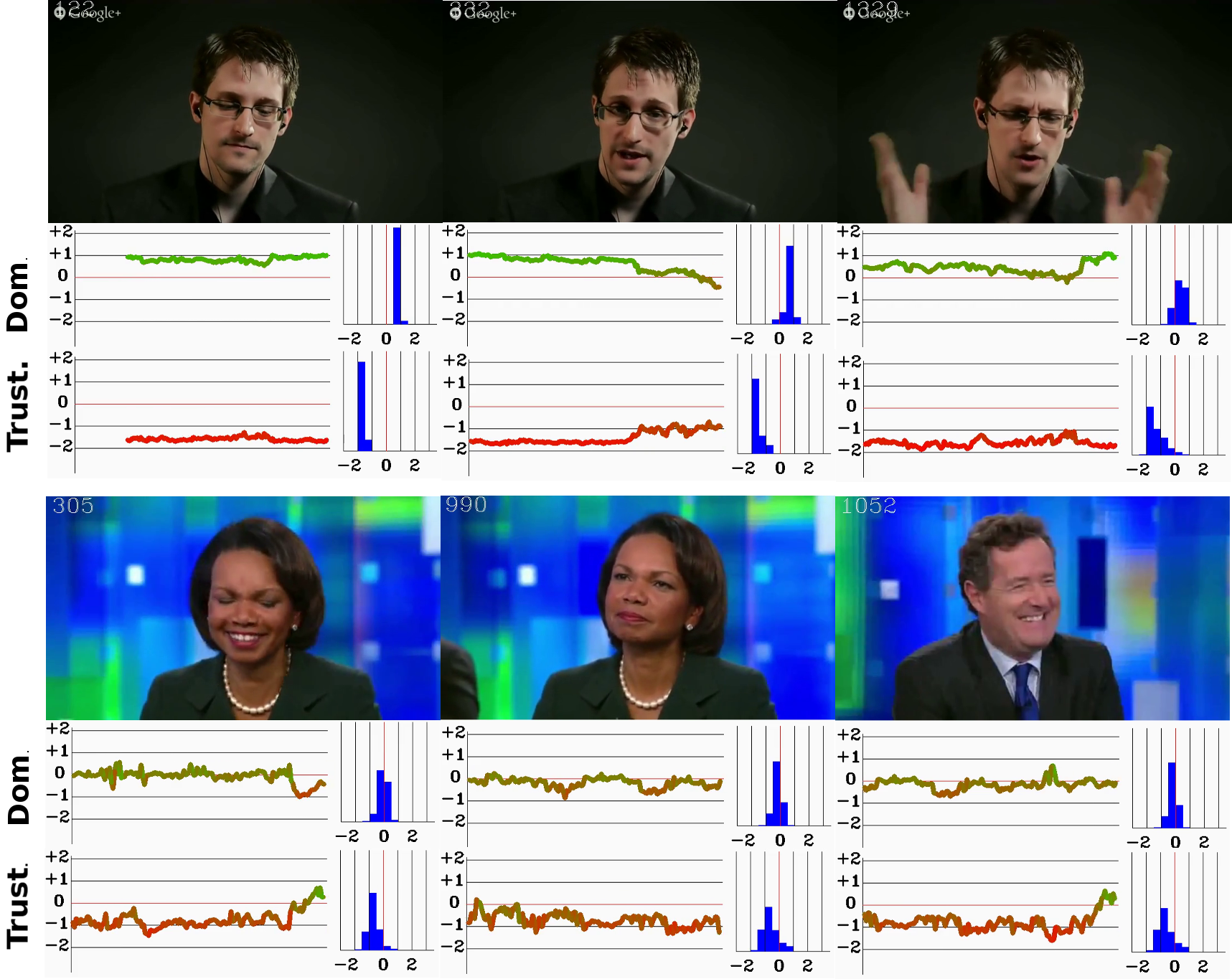}
    \caption{Frames taken from real time video processing examples. The scores are normalized with respect to the training data statistics and then displayed over time on a line plot and a histogram. These frames exemplify changes in predictions based on facial expression and head movement.
    %Similar to Fig.~\ref{fig:teaser}, we attempted to gauge the  trustworthiness and dominance of actors Benedict  Cumberbatch and Joseph Gordon-Levitt, who respectively appear in the movies \textit{The Fifth Estate} and \textit{Snowden}. According to the predictions from the models the actors are rather low on the trustworthiness scale, mirroring the visual consistency observed in Fig.~\ref{fig:teaser}.
    }
    \label{fig:video}
    \vspace{-2mm}
\end{figure*}

\begin{table}[!h]
\begin{center}
\caption{Some important hyperparameter optimization results per trait for our optimized MOON architectures.}
 \label{tab:hyperparameters}
 \begin{tabular}{|| c c c c c ||}
 \hline
  & \textbf{Trust.} & \textbf{Dom.} & \textbf{Age} & \textbf{IQ} \\ [0.3ex] 
 \hline\hline
\textbf{Learning Rate} & $10^{-4.2}$ & $10^{-4.4}$ & $10^{-4.8}$ & $10^{-4.6}$ \\ 
\hline
\textbf{Dropout} & $55\%$ & $31\%$ & $45\%$ & $38\%$ \\
\hline
\textbf{2x Convolution 0} & $64$ & $32$ & $32$ & $64$ \\
\hline
\textbf{2x Convolution 1}  & $64$ & $64$ & $128$ & $32$ \\
\hline
\textbf{2x Convolution 2}  & $128$ & - & - & - \\
\hline
\textbf{3x Convolution 3} & $256$ & $256$ & $256$ & $256$ \\
\hline
\textbf{3x Convolution 4} & $256$ & $512$ & $512$ & $256$ \\ 
\hline
\textbf{3x Convolution 5} & $256$ & $512$ & $512$ & - \\ 
\hline
\textbf{FC Layers} & $1$ & $3$ & $4$ & $3$ \\
\hline
\textbf{FC Outputs} & $2079$ & $2227$ & $2187$ & $1244$ \\
\hline

\end{tabular}
\end{center}
\end{table}

\begin{table}[!h]
\begin{center}
\caption{$R^2$ values of validation and testing results from our optimized MOON architectures for each trait.}
 \label{tab:rsquared}
 \begin{tabular}{|| c c c c c ||}
 \hline
  & \textbf{Trust.} & \textbf{Dom.} & \textbf{Age} & \textbf{IQ} \\ [0.3ex] 
 \hline\hline
\textbf{Validation} & $.41$ & $.49$ & $.75$ & $.29$ \\ 
\hline
\textbf{Test} & $.38$ & $.46$ & $.72$ & $.24$ \\
\hline

\end{tabular}
\end{center}
\end{table}

\subsection{Visualizations of Feature Importance}

Visualizations of the hyperparameter optimized CNN models show localized areas of importance on the face for each trait. As an example, we overlay average heatmaps for each trait on the averaged faces of 100 images from the validation data in Fig.~\ref{fig:averageheatmaps}. To produce these graphics we systematically moved a gray box over an image, iteratively scaling the box down after each pass. We then recorded the absolute difference in total score at each point. This visualization is intriguing because it allows us to view, in a certain image, or over an average of images, what areas of the face have the most or least significant effect on the final prediction. 

Again, it is difficult to assess the validity of our models, as accuracy cannot be calculated because there is no known ground truth. Referring back to previous social psychology research~\cite{mignault2003many, keating1981culture}, however, both Trustworthiness and Dominance are expected to rely on the mouth. Our models indicate a heavy reliance on areas near the mouth and chin. Similarly, Keating et al.~\cite{keating1981culture} determined that a lowered brow should affect the (mostly Western) perception of Dominance. Both our Dominance and Trustworthiness models approximately locate the brow mid-sections. These observations indicate that our models have  learned to look in the same places that humans do, possibly replicating the way we judge high-level attributes in each other. 

Another method of analyzing our models is a visualization of the filters. Our visualizations of the filters from the final convolutional layer of each network in Fig.~\ref{fig:filters} are intriguing because they resemble the output of a low-level feature extractor. This indicates that despite the high-level abstract quality of these traits, low-level features might be enough for humans to make their immediate judgements. This is consistent with our observation that deeper architectures add little to no improvement.

\subsection{Processing Faces in Video}

A very good litmus test for our models is video processing. For each frame from a video, we can apply face detection and face alignment, and then use our optimized models to predict the score of each trait. With the models loaded into memory we can even do this in real time, allowing the subjects in the videos to move and change position, simultaneously determining the change in other people's perceptions. Fig.~\ref{fig:video} shows several frames from a couple of example videos being processed. In Fig.~\ref{fig:video}, all scores are mapped to a standard normal distribution and shown over time on both a line plot and a histogram. A selection of processed videos are provided as supplemental material.  

\section{Discussion}

Current state-of-the-art visual recognition algorithms in computer vision, and more specifically algorithms for facial attribute prediction~\cite{lewenberg2016predicting, rudd2016moon}, show accuracy that promises new applications  in the near future. It is in the best interest of both researchers and developers in industry to promote research that focuses on the interrelation of machine learning, computer vision, and social psychology. 

A model is only as good as its data. The dataset and its annotations will ultimately have the most significant effect on the psychological validity and usefulness of the models. When annotating a dataset for subjective traits, small differences such as the number of annotators, the number of annotations, and the geographic and cultural differences of the annotators must be taken into consideration. Different cultures and languages affect the way people interpret traits, or the description of traits. Just as intriguing as the generalizations about people that we made in our work is the study of different cultures and focus groups. Models trained only on the annotations of a focus group could generalize to new data, enabling cross-culture comparisons --- useful in research, marketing, political campaigning and more.

In systematically analyzing human judgements, it is also important to choose traits that best fulfill a purpose. In our case, Trustworthiness and Dominance are the best representations of the abstract judgements humans make about each other. IQ and Age, while not as fundamental in a psychological sense, still have conceivable applications, including the assessment of preconceived notions of intelligence and seniority --- subtle social cues we often take for granted.

Code, data and supplemental material for this paper can be found at: \url{http://github.com/mel-2445/Predicting-First-Impressions}

\section*{Acknowledgements}
M. McCurrie was supported by a gift from the Boeing Company. F. Beletti and L. Parzianello were supported by the Brazil Scientific Mobility Program. A. Westendorp was supported by NSF CNS RET Award \#1609394. S. Anthony was supported in part by NSF SBIR Award \#IIP-1621689.  Hardware support was generously provided by the NVIDIA Corporation.

{\small
\bibliographystyle{ieee}
\bibliography{main}
}

\end{document}